\documentclass[a4paper, 10pt, conference]{ieeeconf}

\usepackage{colortbl}
\usepackage{amsmath}
\usepackage{wrapfig}
\usepackage{graphicx,stfloats,bm}

\newif\ifarwfinalcopy

\arwfinalcopytrue  

\IEEEoverridecommandlockouts                       

\usepackage{graphics} 
\usepackage{epsfig} 
\usepackage{mathptmx} 
\usepackage{times} 
\usepackage{amsmath} 
\usepackage{amssymb}  

\usepackage{lineno}
\usepackage{tikzpagenodes}
\usepackage{background}

\ifarwfinalcopy
\backgroundsetup{color=white}
\else

\linenumbers

\newcommand{\MyARWConfidentialLogo}{
\begin{tikzpicture}[remember picture,overlay]
\node[align=center,text=blue] at ([yshift=1em]current page text area.north) {\Large \#\#\# ARW 2025 SUBMISSION: CONFIDENTIAL REVIEW COPY \#\#\#};
\end{tikzpicture}%
}

\SetBgContents{\MyARWConfidentialLogo}
\SetBgPosition{current page.north west}
\SetBgOpacity{0.5}
\SetBgAngle{0.0}
\SetBgScale{1.0}

\fi

\title{\LARGE \bf
Multi-Modal 3D Mesh Reconstruction from Images and Text
}

\author{Melvin Reka$^{1}$, Tessa Pulli$^{1}$, and Markus Vincze$^{1}$
\thanks{$^{1}$ all authors are with the Automation and Control Institute, TU Wien
Vienna, Austria: {\tt\small e12102393@student.tuwien.ac.at}; {\tt\small \{pulli, vincze\}@acin.ac.tuwien.at}}}

\begin{document}

\maketitle
\begin{figure*}[bp]  
    \centering
    \includegraphics[width=1\textwidth]{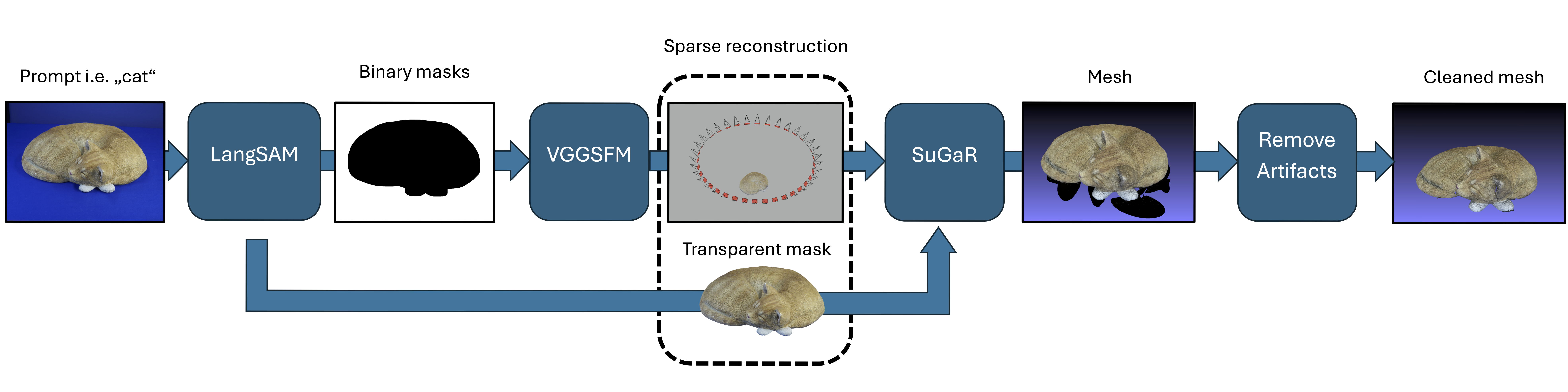}
    \caption{Images of an object, accompanied by a descriptive text prompt, are processed through the pipeline. A sparse reconstruction using COLMAP via VGGSfM is then performed to generate and refine the mesh with SuGaR.}
    \label{fig:pipeline}
\end{figure*}

\begin{abstract}
6D object pose estimation for unseen objects is essential in robotics but traditionally relies on trained models that require large datasets, high computational costs, and struggle to generalize. 
Zero-shot approaches eliminate the need for training but depend on pre-existing 3D object models, which are often impractical to obtain. To address this, we propose a language-guided few-shot 3D reconstruction method, reconstructing a 3D mesh from few input images. 
In the proposed pipeline, receives a set of input images and a language query.
A combination of GroundingDINO and Segment Anything Model outputs segmented masks from which a sparse point cloud is reconstructed with VGGSfM. 
Subsequently, the mesh is reconstructed with the Gaussian Splatting method SuGAR.
In a final cleaning step, artifacts are removed, resulting in the final 3D mesh of the queried object.
We evaluate the method in terms of accuracy and quality of the geometry and texture.
Furthermore, we study the impact of imaging conditions such as viewing angle, number of input images, and image overlap on 3D object reconstruction quality, efficiency, and computational scalability.
 
\end{abstract}

\begin{keywords}
Vision Language Models, Language-guided Reconstruction, Few-shot Reconstruction
\end{keywords}

\section{Introduction}
6D object pose estimation for unseen objects is a critical task in robotics.
Traditional methods estimate instance object poses using trained networks~\cite{wang2021gdr, su2022zebrapose, park2019pix2pose, xiang2017posecnn}.
However, training models for object pose estimation is a limitation as it requires large annotated datasets, has high computational costs, and encounters difficulties in generalizing to unknown objects or environments~\cite{bauer2024challenges}.
An alternative to these methods are training-free zero-shot approaches~\cite{ausserlechner2024zs6d, thalhammer2023self}.
Methods such as ZS6D~\cite{ausserlechner2024zs6d} utilize a ground truth object model to find 2D-3D correspondences between the model and the images from which the pose is computed using a PnP algorithm~\cite{hönig2024improving2d3ddensecorrespondences}. 
Zero-shot object pose estimation algorithms offer a considerable advantage because they do not require a training phase.
However, these methods are limited by the requirement for a ground-truth 3D object model\cite{ausserlechner2024zs6d, su2022zebrapose}, meaning that they depend on prior knowledge of objects encountered during inference. 
This poses a challenge, as obtaining high-quality 3D models can be labor intensive, expensive, and impractical for large-scale or real-time applications\cite{shen2023anything}.
With the advent of diffusion models, recent works have proposed methods for a few-image\cite{qian2023magic123, guédon2023sugarsurfacealignedgaussiansplatting} or even single-shot~\cite{qian2023magic123, long2024wonder3d} 3D model reconstruction based on images.
By combining SuGAR~\cite{guédon2023sugarsurfacealignedgaussiansplatting} with SAM~\cite{kirillov2023segany} and GroundingDINO~\cite{liu2023grounding}, we introduce a novel method for reconstructing 3D models based on images and language prompts.

We propose a language-guided few-shot reconstruction method. 
As input, we receive several RGB images of a scene. According to the language input, the queried object is masked with a padding of 50 pixels with a combination of SAM~\cite{kirillov2023segany} and GroundingDINO~\cite{liu2023grounding}.
The masked images are then used as input for sparse reconstruction with VGGSfM~\cite{wang2023visualgeometrygroundeddeep}, which produces the sparse reconstruction for further mesh generation with SuGAR~\cite{guédon2023sugarsurfacealignedgaussiansplatting}. 
Finally, the reconstruction is evaluated on several experiments.

In summary, the paper has the following
key contributions:
\begin{itemize}
    \item We propose a novel language-guided few-shot reconstruction approach that allows 3D model reconstruction.
    \item Evaluation of the few-shot reconstruction method, including an analysis of the required input images considering efficiency and performance of the reconstruction.
\end{itemize}

The rest of this work is organized as follows: Section II introduces the related work of 6D object pose estimation, 3D model reconstruction, and language guided segmentation.
Section III describes the pipeline of our proposed methods.
In Section IV, the experimental setup is presented while Section V discusses the evaluation. Section VI concludes the paper with a summary and outlook.
\section{Related work}
In this section, we discuss the related work by revisiting 6D object pose estimation, few-shot reconstruction methods and language guided object segmentation.

\subsection{6D Object Pose Estimation}
6D object pose estimation of unseen objects is a core task in robotics. 
Classical methods estimate the pose of objects using trained networks for object instances~\cite{wang2021gdr, su2022zebrapose, park2019pix2pose}.
These methods require large annotated datasets, which are costly to acquire while requiring high computational power during training~\cite{bauer2024challenges}. 
Furthermore, these models struggle to generalize to unseen objects~\cite{wang2021gdr, park2019pix2pose, xiang2017posecnn}, limiting their applicability in real-world scenarios where those objects frequently appear.
These challenges can be overcome with zero-shot object pose estimation methods~\cite{ausserlechner2024zs6d, thalhammer2023self}.
These approaches eliminate the need for extensive training by leveraging prior knowledge by assuming that object models exist.
Given the reference model, 2D-3D correspondences between the object's model and a set of input images can be established~\cite{hönig2024improving2d3ddensecorrespondences}.
With this information available, the object pose is computed by a PnP/RANSAC algorithm~\cite{li2023nerf}.
While zero-shot methods offer a significant advantage by overcoming the training phase, they still require high-quality ground-truth 3D models. 
This dependency presents challenges in practical applications, as obtaining accurate 3D models is labor-intensive, requiring expensive equipment~\cite{choi2023tmo}.

\subsection{Few-Shot 3D Reconstruction Methods}
Recent advances in generative models, particularly diffusion models, have opened new possibilities to acquire 3D meshes.
Works such as Wonder3D~\cite{long2024wonder3d}, Gaussian Surfels~\cite{Dai2024GaussianSurfels}, Dreamfusion~\cite{poole2022dreamfusion}, and Sugar~\cite{guédon2023sugarsurfacealignedgaussiansplatting} have demonstrated the potential of generating 3D object representations from a limited number of 2D images, enabling object reconstruction without requiring pre-existing CAD models.
Typically, these methods use 2D diffusion models to generate novel views from a different camera viewpoint with a diffusion model.
From these novel views a 3D model is reconstructed with a stochastic 3D reconstruction framework~\cite {poole2022dreamfusion} or can be parameterized as a voxel radiance field from which the mesh is extracted with a marching cubes procedure~\cite{liu2023zero}. 
Because existing reconstruction methods typically assume clean, isolated object inputs, several methods introduced a pre-processing step to delete the background of the input images to avoid noise while reconstructing the images~\cite{xu2024instantmesh}. 

\subsection{Language Guided Object Segmentation}
The integration of vision-language models (VLMs) such as CLIP~\cite{clip} has significantly expanded the capabilities of computer vision systems, enabling them to understand and process images based on textual descriptions.
CLIPP~\cite{clip} has been incorporated into a wide range of applications allowing methods including scene understanding~\cite{chen2023clip2scene}, object recognition~\cite{zhang2024recognize}, and generative modeling~\cite{poole2022dreamfusion}, offering new possibilities for interactive vision-based tasks. 
An area where this integration has proven particularly beneficial is object segmentation~\cite{kirillov2023segany}, where language-guided approaches allow for more intuitive and adaptable object selection.
Recent advancements in segmentation models, particularly the Segment Anything Model (SAM), provide a method to segment objects in an image using minimal user input. 
SAM allows for object selection through different means such as points, bounding boxes,  or language-based prompts, making it an effective tool for isolating objects in complex scenes. 
By leveraging SAM’s capabilities, objects can be accurately segmented and masked before being passed into a 3D reconstruction pipeline.

\section{Method}
As shown in Figure \ref{fig:pipeline}, images of the object, along with a descriptive text prompt, are processed through the pipeline. 
LangSAM \cite{medeiros2023langsegmentanything} combines GroundingDINO \cite{liu2023grounding} and SAM \cite{kirillov2023segany} for text-driven segmentation. 
GroundingDINO processes the text prompt to generate bounding boxes around relevant objects, which SAM then uses to create binary masks with a 50-pixel paddeing. 
This padding proved helpful when handling semi-transparent objects and reduces artifacts introduced during mesh reconstruction. 
By focusing on essential scan areas, this approach enhances reconstruction quality while maintaining computational efficiency. \\
The generated masks, along with the original images, are then used for 3D reconstruction with VGGSfM \cite{wang2023visualgeometrygroundeddeep}. 
As a fully differentiable structure-from-motion pipeline, VGGSfM estimates camera parameters, determines camera positions, and reconstructs a sparse point cloud by tracking corresponding 2D points across multiple views. 
This end-to-end differentiable approach enhances the accuracy and robustness of the reconstruction process by eliminating the need for chaining pairwise matches and enabling simultaneous recovery these\cite{wang2023visualgeometrygroundeddeep}. \\
Using the resulting COLMAP\cite{schoenberger2016sfm} dataset and the extracted RGB masks, a textured mesh is generated with SuGaR\cite{guédon2023sugarsurfacealignedgaussiansplatting}, which employs Gaussian Splatting to efficiently optimize and extract a high-resolution 3D surface. 
However, since SuGaR can introduce artifacts during mesh generation, an automated script utilizing PyMeshLab\cite{pymeshlab} is applied to remove these artifacts, ensuring a cleaner final reconstruction. 

\section{Experiments} \label{experiments}
In this section, the experimental setup is discussed.
During our evaluation, we assess the accuracy and quality of the geometric construction as well as the reconstructed texture. 

\subsection{Implementation details:} 
All experiments are conducted on a system equipped with an AMD Ryzen 9 5950X CPU, 128GB RAM, and an NVIDIA RTX 3090 GPU with 24GB VRAM. 
The implementation is containerized using Docker to ensure reproducibility across different hardware environments.

\subsection{Evaluation Metrics}
To assess the accuracy and quality of both geometric reconstruction and texture extraction, we distinguish between geometric metrics and 
texture similarity. 
\vspace{-1pt}
\subsection*{\textbf{Geometric Metrics}}
\vspace{-1pt}
We evaluate the reconstructed 3D geometry using Chamfer Distance~\cite{point-cloud-utils} and Intersection over Union~\cite{Wang-SIG2022}. 
CD quantifies the average squared distance between nearest neighbors in the predicted and ground truth meshes.
It is defined as:
\begin{equation}
CD(P, Q) = \frac{1}{|P|} \sum_{p \in P} \min_{q \in Q} \|p - q\|^2 + \frac{1}{|Q|} \sum_{q \in Q} \min_{p \in P} \|q - p\|^2
\end{equation}
where \( P \) and \( Q \) are the sets of points in the predicted and ground truth meshes. 
Low CD values indicate a more precise geometric alignment.

IoU measures the volumetric similarity between the reconstructed and ground truth models:
\begin{equation}
IoU = \frac{|V_P \cap V_Q|}{|V_P \cup V_Q|}
\end{equation}
where \( V_P \) and \( V_Q \) represent the volumetric reconstructions of the predicted and ground truth meshes. 
IoU measures the proportion of the shared volume, with higher values indicating better alignment. 
\vspace{-1pt}
\subsection*{\textbf{Texture Similarity}}
\vspace{-1pt}
To assess texture extraction accuracy, we use the three key metrics employed in SuGaR~\cite{guédon2023sugarsurfacealignedgaussiansplatting}.\\
The Peak Signal-to-Noise Ratio is defined as:
\begin{equation}
PSNR = 10 \log_{10} \left(\frac{MAX^2}{MSE} \right)
\end{equation}
where \( MAX \) is the maximum possible pixel value, and \( MSE \) is the mean squared error between the predicted and ground truth textures. 
Higher values indicate better pixel-wise preservation, but do not reflect human perception.\\

The Structural Similarity Index~\cite{1284395} takes into account luminance, contrast, and texture integrity, reflecting human perception.
SSIM is computed as:
\begin{equation}
SSIM(x, y) = \frac{(2\mu_x\mu_y + C_1)(2\sigma_{xy} + C_2)}{(\mu_x^2 + \mu_y^2 + C_1)(\sigma_x^2 + \sigma_y^2 + C_2)}
\end{equation}
where \( \mu_x \) and \( \mu_y \) are the mean intensities, \( \sigma_x^2 \) and \( \sigma_y^2 \) are the variances, and \( \sigma_{xy} \) is the covariance between the predicted and ground truth images. \\

The Learned Perceptual Image Patch Similarity~\cite{zhang2018perceptual} is given by:
\begin{equation}
LPIPS(x, y) = \sum_l w_l  \left\| F_l(x) - F_l(y) \right\|_2^2
\end{equation}
where \( F_l \) represents the feature maps at layer \( l \) of a pretrained network, and \( w_l \) are learned weights. LPIPS captures high-level perceptual differences, making it effective for identifying distortions and artifacts beyond pixel-wise comparisons.

\subsection{Experimental Setup}
Our experiments evaluate the impact of various imaging conditions on the quality of 3D object reconstruction. 
We investigate how the viewing angle \(\theta\) influences feature extraction and reconstruction accuracy, as different angles affect feature visibility. 
We also examine the effect of the number of input images on reconstruction convergence, assessing how multi-view stereo improves model quality. 
The overlap between input images, determined by rotation step sizes \(\phi\), is another factor influencing reconstruction accuracy. 
Mesh quality is assessed by comparing the texture extraction accuracy and alignment with ground truth. 
Finally, a runtime analysis measures the scalability of computational costs with the number of input images and processing steps, balancing efficiency and accuracy in real-time applications. 
These experiments aim to understand the influence of these parameters on reconstruction quality, efficiency, and robustness.

\begin{figure}[h] 
  \centering
  \includegraphics[width=0.5\linewidth]{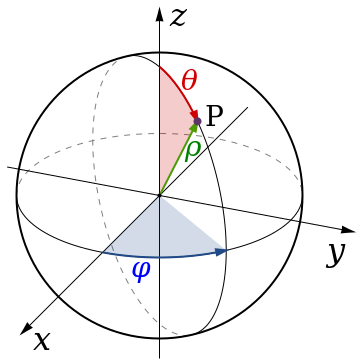}
  \caption{Spherical coordinates, where we refer to the polar angle $\theta$ as viewing angle, and to the azimuthal angle $\phi$ as }
  \label{fig:spherical}
\end{figure}

Furthermore, we assess mesh quality, focusing on the accuracy of texture extraction and the overall fidelity of the reconstructed surfaces. Finally, we conduct a runtime analysis to measure how computational cost scales with the number of input images and processing steps, balancing efficiency and accuracy in multi-view stereo reconstruction.

\subsection{Dataset}

To evaluate the capabilities of the proposed work, we utilize the MVS dataset~\cite{Shuji_SAKAI20152014EDP7409} consisting of two sets of multi-view images, their camera parameters, and the ground-truth mesh models. 
It includes multiple views of each scene, captured from different angles to provide a diverse set of perspectives. 
The viewing angles are characterized by two key parameters: theta ($\theta$), the polar angle, which defines the elevation or vertical angle from which the scene is viewed.
\ref{fig:spherical}.
\[
\theta_{\text{cat}} \in \{30^\circ, 45^\circ, 75^\circ\}
\wedge
\theta_{\text{dog}} \in \{ 45^\circ, 90^\circ\}
\]

\begin{figure}[h] 
    \centering
    \includegraphics[width=1\linewidth]{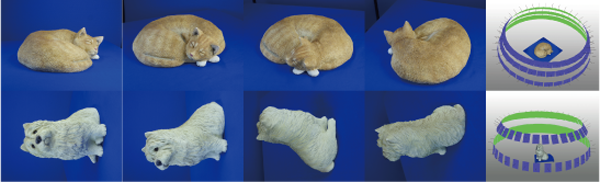}
    \caption{THU Multi-view stereo datasets~\cite{Shuji_SAKAI20152014EDP7409} of a cat and a dog. The left side shows the input images captured from various viewpoints, while the right side displays the corresponding camera viewpoints and the target objects.}
    \label{fig:dataset_example}
\end{figure}

Phi $\phi$, the azimuthal angle, which represents the horizontal rotation around the scene. 
By capturing images across varying $\theta$ and $\phi$ angles, the dataset offers a comprehensive range of viewpoints. 
Figure \ref{fig:dataset_example} shows examples of the dataset and its target objects which are figurines of a cat and a dog. 
The images are taken with the camera by changing the height of the camera with three different viewing angles. 
$\theta$ is the polar angle between the z-axis and the camera position, which we refer to as $\theta$, as it can be seen in Figure \ref{fig:dataset_example}.

\section{Evaluation} \label{evaluation}
This evaluation assesses the performance of the proposed method based on input images, focusing on the trade-off between accuracy and efficiency. 
Specifically, we analyze which input configurations yield the most precise reconstructions while maintaining computational efficiency.

\subsection*{\textbf{Number of Input Images}}
\begin{figure} [h]
    \centering
    \includegraphics[width=1\linewidth]{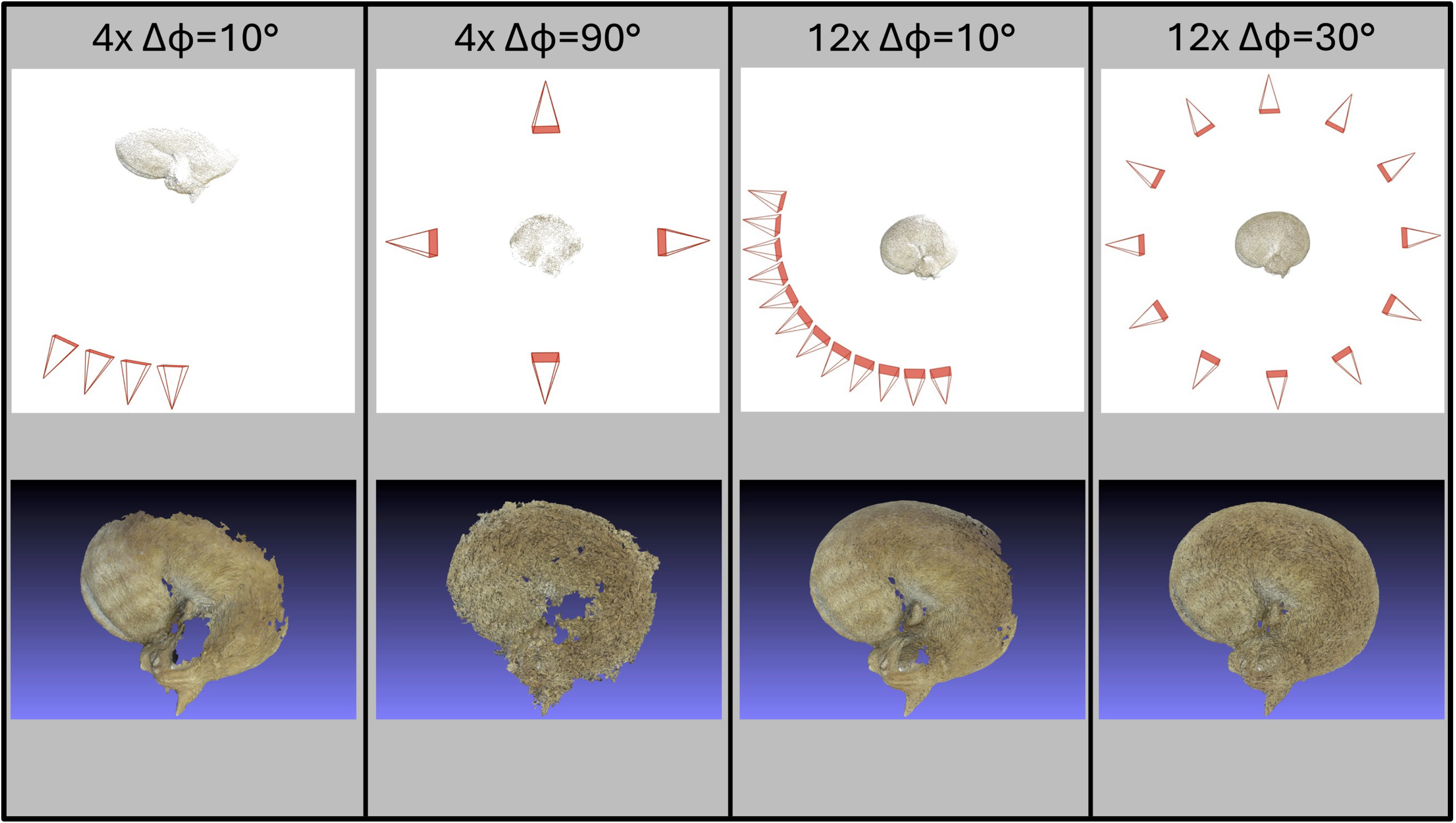}
    \caption{Overlap of input images comparison for $\theta=45^\circ$}
    \label{fig:number_of_images}
\end{figure}

In a first experiment, we show the impact of the number of input images on the geometric reconstruction quality.
We used three different sets of images according to the three camera viewing angle $\theta$ (30°, 45°, 75°). 
While several combinations of rotation angles between images are possible, we chose the best result for each number of images, neglecting factors such as overlap between images as these are investigated in the following experiments.
Figute~\ref{fig:number_of_images} shows that for each of these data sets, the model converges for both CD and IoU  approximately from 15 images onward while the best performance is achieved with the maximum number of input images of 36.
However, some outliers deviate significantly from this trend, which can be attributed to the effects of overlap and coverage.

\subsection*{\textbf{Runtime}}
\begin{figure} [h]
    \centering
    \includegraphics[width=1\linewidth]{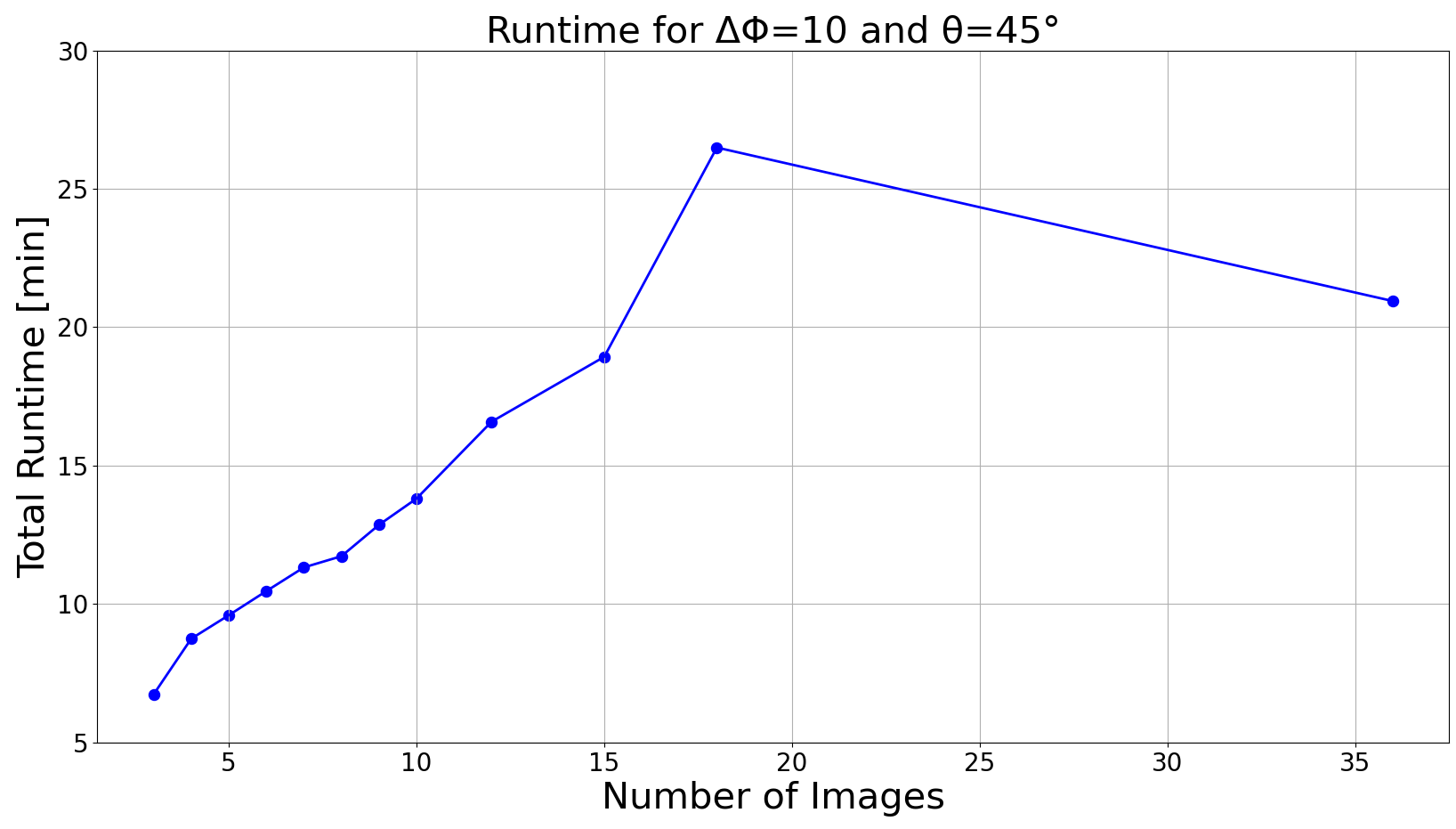}
    \caption{Runtime vs. input images for \(\theta = 45^\circ, \Delta\phi = 10^\circ\)}
    \label{fig:runtime}
\end{figure}

In Figure \ref{fig:runtime}, the model exhibits a linear runtime increase with the number of images up to 18. At 36 images, a drop-off occurs as VGGSfM is downscaled by half to fit within the available VRAM, resulting in a lower-quality reconstruction but a more accurate overall outcome. The runtime is divided into three parts: segmentation (negligible), sparse reconstruction (scales linearly with image count), and mesh extraction (consistently 4–5 minutes).

\subsection*{\textbf{Viewing Angle Theta $\theta$}} 
\begin{figure}[h]
    \centering
    \includegraphics[width=1\linewidth]{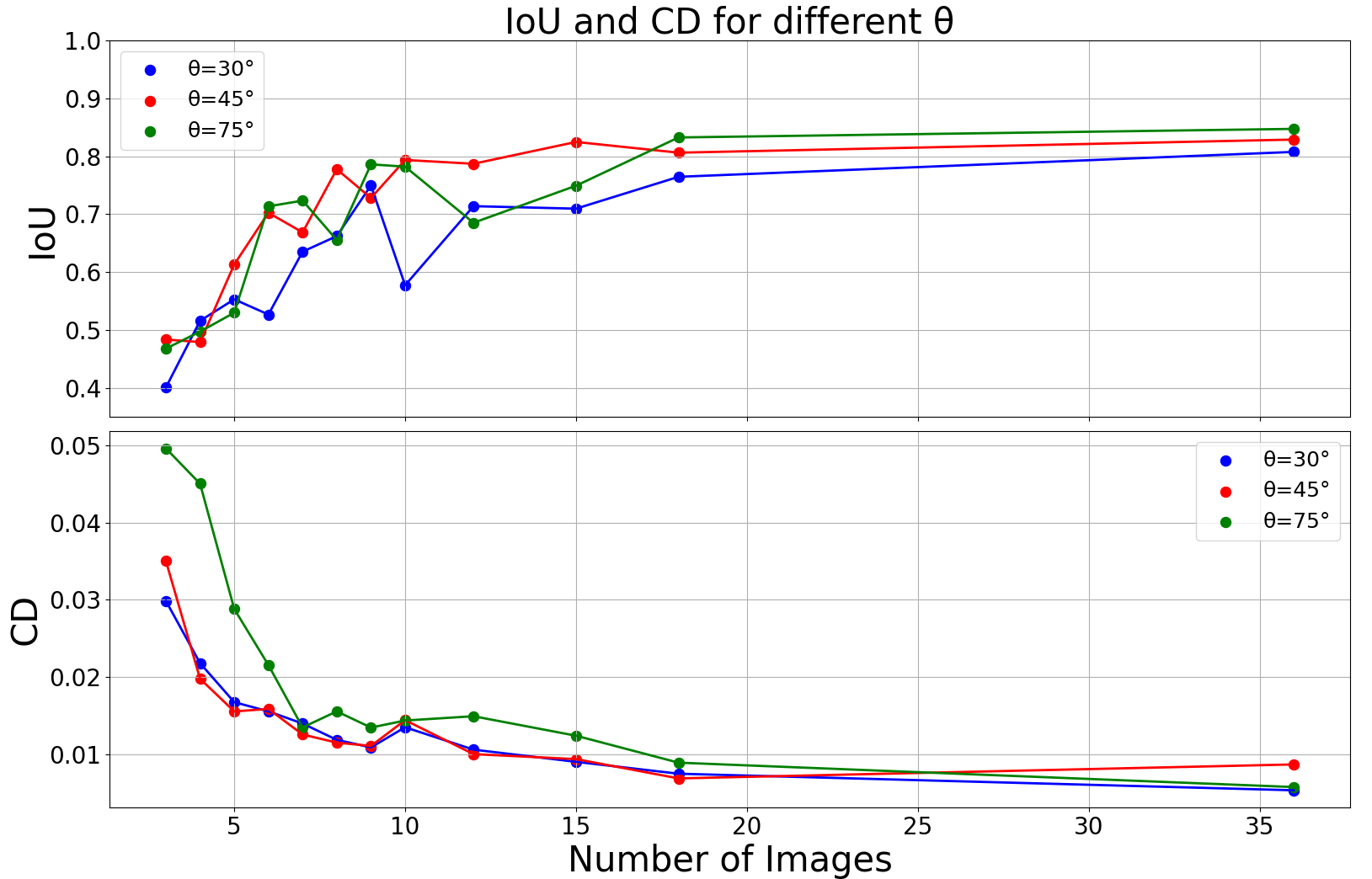}
    \caption{IoU and Chamfer Distance for cat figurine at different \(\theta\) angles.}
    \label{fig:camer_angles}
\end{figure}

In the first experiment, we investigate how different camera angles $\theta$ affect reconstruction quality.
Therefore, the relationship between the number of images and different camera angles is used on the example of the cat and dog figurines.
Figure \ref{fig:camer_angles} presents the best reconstruction results in terms of IoU and CD for each tested viewing angle $\theta$.
While multiple configurations are possible, we report only the optimal results for each angle.

Our findings indicate that the ideal angle of incidence for both objects is 45°. 
At this angle, VGGSfM achieves the most effective feature extraction, as it captures both the top and front of the object within the same image. 
This results in a higher number of extracted points without increasing the total number of input images, outperforming alternative angles such as 30° and 75° for the cat and 90° for the dog.

\begin{figure} [h]
    \centering
    \includegraphics[width=1\linewidth]{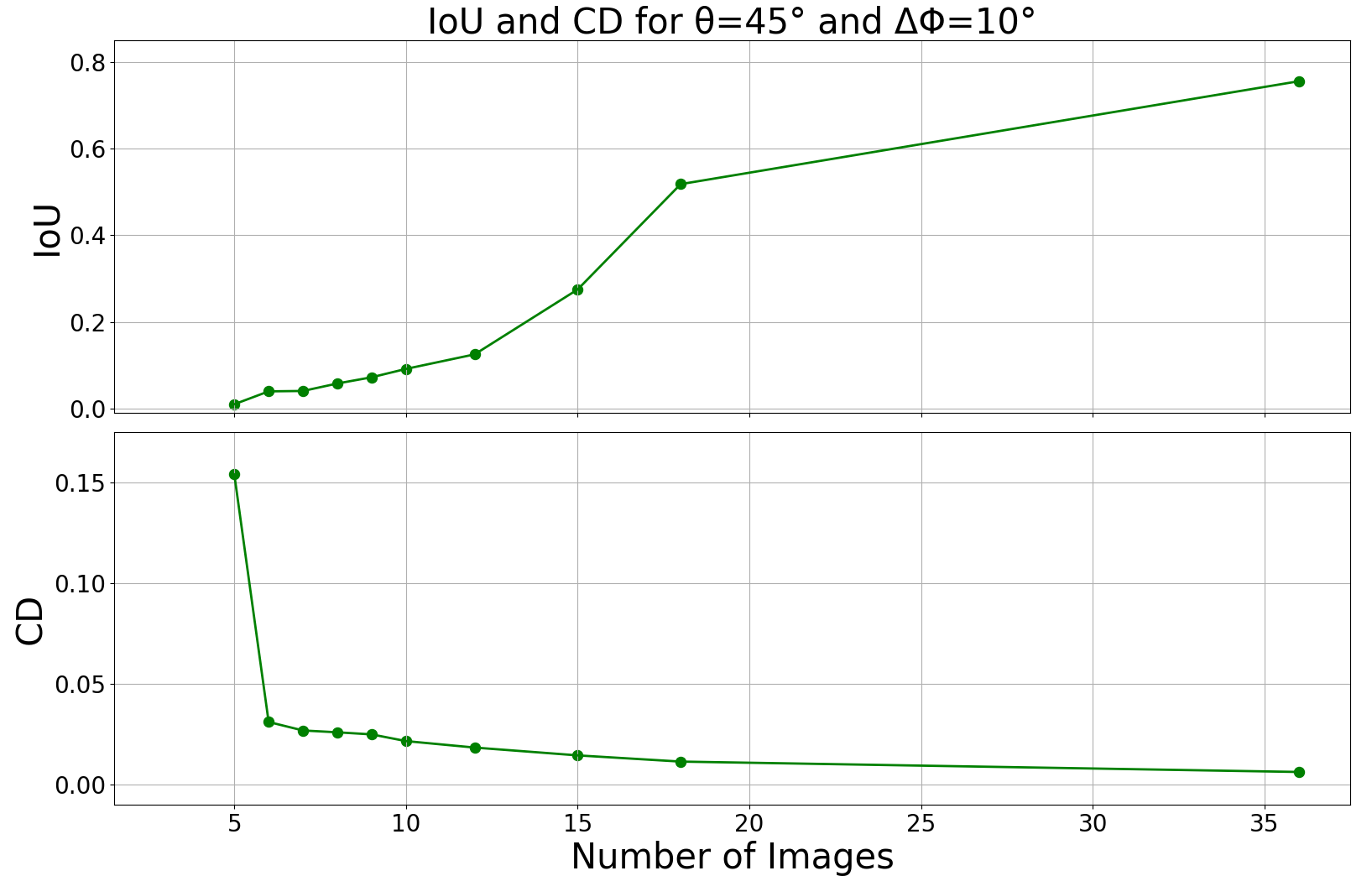}
    \caption{IoU and Chamfer Distance for dog figurine at fixed $\theta$ and $\Delta\phi$}
    \label{fig:theta_phi}
\end{figure}

\subsection*{\textbf{Overlap of Input Images}}
To reduce runtime and the computational effort required for mesh reconstruction, the following experiments explore how the input image set can be optimized.
A key factor in this process is the overlap of input images. 
With a small rotation step (\(\Delta\phi\)) and few images, the visible areas are well-reconstructed, but limited coverage lowers benchmark scores despite reasonable results for small datasets. 
On the other hand, using few images and a large rotation step  (\(\Delta\phi\)), leads to insufficient feature overlap and therefore an incomplete mesh. 
Increasing the number of images with a small \(\Delta\phi\) enhances reconstruction quality, but the results are still limited to the visible areas. 
A balanced approach, such as 12 images with \(\Delta\phi = 30^\circ\)), achieves comparable quality to using three times as many images, improving both accuracy and coverage.

\begin{table}[h]
  \centering
  \resizebox{\linewidth}{!}{ 
  \begin{tabular}{|c|c|c|c|c|}
    \hline
    \textbf{Nr. of images} & \textbf{$\Delta\phi$ [DEG]} & \textbf{CD $\downarrow$} & \textbf{IoU [\%] $\uparrow$} & \textbf{Runtime [min]} \textbf{$\downarrow$} \\ \hline
    4   & 10°   & 0.0376 & 47.91 & \textbf{8.75}   \\ \hline
    4   & 90°   & 0.0277 & 43.52 & 9.70 \\ \hline
    12  & 10°   & 0.0162 & 63.32 & 16.59  \\ \hline
    12  & 30°   & 0.0100 & 78.72 & 16.06 \\ \hline
    36  & 10°   & \textbf{0.0087} & \textbf{82.91} & 21.95 \\ \hline
  \end{tabular}
  }
  \caption{Overlap extremes vs. image count for cat at \(\theta = 45^\circ\)}
  \label{tab:overlaps}
\end{table}

\subsection*{\textbf{Mesh Quality}} 
As shown in Table \ref{tab:Mesh_quality}, images taken with smaller $\theta$ angles, such as top-down views, result in better texture similarities. 
Top-down perspectives provide a clearer view of surface details, making it easier to capture fine-grained textures.
In contrast, frontal views with higher $\theta$ values lead to lower quality meshes due to occlusions and the lack of sufficient top-view coverage, making it difficult to accurately reconstruct the surface.
Optimizing the pipeline and capturing images from the optimal angle leads to significantly improved SSIM and LPIPS scores, which, in human perception, translates to textures that closely resemble the ground truth.

\begin{table}[h]
  \centering 
  \begin{tabular}{|c|c|c|c|c|}
    \hline
    \textbf{Figurine} & \textbf{$\theta$ [DEG]} & \textbf{PSNR [dB] $\uparrow$} & \textbf{SSIM $\uparrow$} & \textbf{LPIPS $\downarrow$}  \\ \hline
    Cat   & 30°   & \textbf{33.95}  & \textbf{0.967} & \textbf{0.0373}  \\ \hline
    Cat   & 45°   & 31.75  & 0.947 & 0.0546  \\ \hline
    Cat   & 75°   & 13.52  & 0.644  & 0.2516   \\ \hline
    Dog   & 45°   & \textbf{34.35}  & \textbf{0.972} & \textbf{0.0353}   \\ \hline
    Dog   & 90°   & 22.66  & 0.843  & 0.1195   \\ \hline
    \hline
    \rowcolor{gray!20}  
    \textbf{SuGaR} &  /  & 29.43  & 0.910 & 0.216  \\ \hline
  \end{tabular}
  \caption{Texture quality at different $\theta$ with 36 images. Values printed in \textbf{bold} indicate the best result for each figurine}
  \label{tab:Mesh_quality}
\end{table}


\section{SUMMARY AND OUTLOOK}
This paper analyzes how the angle of incidence and the angular distance between input images affect photogrammetric reconstruction quality.
We show that an optimal balance between image count and angular separation significantly enhances mesh quality, while excessive gaps hinder feature matching.
A key challenge identified is scale estimation, which could be improved by integrating a reference object for automatic scaling. Additionally, sparse reconstruction is a major computational bottleneck, suggesting the need for more efficient alternatives. Future work should focus on optimizing reconstruction pipelines to improve runtime and scale consistency, making the approach more suitable for large-scale or real-time applications.

\addtolength{\textheight}{-12cm}   

\section*{ACKNOWLEDGMENT}

We gratefully acknowledge the support of the EU-program EC Horizon 2020 for Research and
Innovation under project No. I 6114, project iChores.

{\small
\bibliographystyle{IEEEtranS}
\bibliography{bib.bib}

\begin{thebibliography}{10}
\providecommand{\url}[1]{#1}
\csname url@rmstyle\endcsname
\providecommand{\newblock}{\relax}
\providecommand{\bibinfo}[2]{#2}
\providecommand\BIBentrySTDinterwordspacing{\spaceskip=0pt\relax}
\providecommand\BIBentryALTinterwordstretchfactor{4}
\providecommand\BIBentryALTinterwordspacing{\spaceskip=\fontdimen2\font plus
\BIBentryALTinterwordstretchfactor\fontdimen3\font minus \fontdimen4\font\relax}
\providecommand\BIBforeignlanguage[2]{{%
\expandafter\ifx\csname l@#1\endcsname\relax
\typeout{** WARNING: IEEEtran.bst: No hyphenation pattern has been}%
\typeout{** loaded for the language `#1'. Using the pattern for}%
\typeout{** the default language instead.}%
\else
\language=\csname l@#1\endcsname
\fi
#2}}

\bibitem{ausserlechner2024zs6d}
P.~Ausserlechner, D.~Haberger, S.~Thalhammer, J.-B. Weibel, and M.~Vincze, ``Zs6d: Zero-shot 6d object pose estimation using vision transformers,'' in \emph{2024 IEEE International Conference on Robotics and Automation (ICRA)}.\hskip 1em plus 0.5em minus 0.4em\relax IEEE, 2024, pp. 463--469.

\bibitem{bauer2024challenges}
D.~Bauer, P.~H{\"o}nig, J.-B. Weibel, J.~Garc{\'\i}a-Rodr{\'\i}guez, M.~Vincze, \emph{et~al.}, ``Challenges for monocular 6d object pose estimation in robotics,'' \emph{IEEE Transactions on Robotics}, 2024.

\bibitem{chen2023clip2scene}
R.~Chen, Y.~Liu, L.~Kong, X.~Zhu, Y.~Ma, Y.~Li, Y.~Hou, Y.~Qiao, and W.~Wang, ``Clip2scene: Towards label-efficient 3d scene understanding by clip,'' in \emph{Proceedings of the IEEE/CVF Conference on Computer Vision and Pattern Recognition}, 2023, pp. 7020--7030.

\bibitem{choi2023tmo}
J.~Choi, D.~Jung, T.~Lee, S.~Kim, Y.~Jung, D.~Manocha, and D.~Lee, ``Tmo: Textured mesh acquisition of objects with a mobile device by using differentiable rendering,'' in \emph{Proceedings of the IEEE/CVF Conference on Computer Vision and Pattern Recognition}, 2023, pp. 16\,674--16\,684.

\bibitem{Dai2024GaussianSurfels}
P.~Dai, J.~Xu, W.~Xie, X.~Liu, H.~Wang, and W.~Xu, ``High-quality surface reconstruction using gaussian surfels,'' in \emph{ACM SIGGRAPH 2024 Conference Papers}.\hskip 1em plus 0.5em minus 0.4em\relax Association for Computing Machinery, 2024.

\bibitem{guédon2023sugarsurfacealignedgaussiansplatting}
\BIBentryALTinterwordspacing
A.~Guédon and V.~Lepetit, ``Sugar: Surface-aligned gaussian splatting for efficient 3d mesh reconstruction and high-quality mesh rendering,'' 2023. [Online]. Available: \url{https://arxiv.org/abs/2311.12775}
\BIBentrySTDinterwordspacing

\bibitem{hönig2024improving2d3ddensecorrespondences}
\BIBentryALTinterwordspacing
P.~Hönig, S.~Thalhammer, and M.~Vincze, ``Improving 2d-3d dense correspondences with diffusion models for 6d object pose estimation,'' 2024. [Online]. Available: \url{https://arxiv.org/abs/2402.06436}
\BIBentrySTDinterwordspacing

\bibitem{kirillov2023segany}
A.~Kirillov, E.~Mintun, N.~Ravi, H.~Mao, C.~Rolland, L.~Gustafson, T.~Xiao, S.~Whitehead, A.~C. Berg, W.-Y. Lo, P.~Doll{\'a}r, and R.~Girshick, ``Segment anything,'' \emph{arXiv:2304.02643}, 2023.

\bibitem{li2023nerf}
F.~Li, S.~R. Vutukur, H.~Yu, I.~Shugurov, B.~Busam, S.~Yang, and S.~Ilic, ``Nerf-pose: A first-reconstruct-then-regress approach for weakly-supervised 6d object pose estimation,'' in \emph{Proceedings of the IEEE/CVF International Conference on Computer Vision}, 2023, pp. 2123--2133.

\bibitem{liu2023zero}
R.~Liu, R.~Wu, B.~Van~Hoorick, P.~Tokmakov, S.~Zakharov, and C.~Vondrick, ``Zero-1-to-3: Zero-shot one image to 3d object,'' in \emph{Proceedings of the IEEE/CVF international conference on computer vision}, 2023, pp. 9298--9309.

\bibitem{liu2023grounding}
S.~Liu, Z.~Zeng, T.~Ren, F.~Li, H.~Zhang, J.~Yang, C.~Li, J.~Yang, H.~Su, J.~Zhu, \emph{et~al.}, ``Grounding dino: Marrying dino with grounded pre-training for open-set object detection,'' \emph{arXiv preprint arXiv:2303.05499}, 2023.

\bibitem{long2024wonder3d}
X.~Long, Y.-C. Guo, C.~Lin, Y.~Liu, Z.~Dou, L.~Liu, Y.~Ma, S.-H. Zhang, M.~Habermann, C.~Theobalt, \emph{et~al.}, ``Wonder3d: Single image to 3d using cross-domain diffusion,'' in \emph{Proceedings of the IEEE/CVF conference on computer vision and pattern recognition}, 2024, pp. 9970--9980.

\bibitem{medeiros2023langsegmentanything}
L.~Medeiros, ``Langsegmentanything: A repository for segmentation using language models,'' \url{https://github.com/luca-medeiros/lang-segment-anything/tree/0a12766ced0503f16ee50e6fa99a9a4f5fbd4ea5}, 2023, accessed: 2025-02-08.

\bibitem{pymeshlab}
A.~Muntoni and P.~Cignoni, ``{PyMeshLab},'' Jan. 2021.

\bibitem{park2019pix2pose}
K.~Park, T.~Patten, and M.~Vincze, ``Pix2pose: Pixel-wise coordinate regression of objects for 6d pose estimation,'' in \emph{Proceedings of the IEEE/CVF International Conference on Computer Vision}, 2019, pp. 7668--7677.

\bibitem{poole2022dreamfusion}
B.~Poole, A.~Jain, J.~T. Barron, and B.~Mildenhall, ``Dreamfusion: Text-to-3d using 2d diffusion,'' \emph{arXiv preprint arXiv:2209.14988}, 2022.

\bibitem{qian2023magic123}
G.~Qian, J.~Mai, A.~Hamdi, J.~Ren, A.~Siarohin, B.~Li, H.-Y. Lee, I.~Skorokhodov, P.~Wonka, S.~Tulyakov, \emph{et~al.}, ``Magic123: One image to high-quality 3d object generation using both 2d and 3d diffusion priors,'' \emph{arXiv preprint arXiv:2306.17843}, 2023.

\bibitem{clip}
\BIBentryALTinterwordspacing
A.~Radford, J.~W. Kim, C.~Hallacy, A.~Ramesh, G.~Goh, S.~Agarwal, G.~Sastry, A.~Askell, P.~Mishkin, J.~Clark, G.~Krueger, and I.~Sutskever, ``Learning transferable visual models from natural language supervision,'' 2021. [Online]. Available: \url{https://arxiv.org/abs/2103.00020}
\BIBentrySTDinterwordspacing

\bibitem{Shuji_SAKAI20152014EDP7409}
S.~SAKAI, K.~ITO, T.~AOKI, T.~WATANABE, and H.~UNTEN, ``Phase-based window matching with geometric correction for multi-view stereo,'' \emph{IEICE Transactions on Information and Systems}, vol. E98.D, no.~10, pp. 1818--1828, 2015.

\bibitem{schoenberger2016sfm}
J.~L. Sch\"{o}nberger and J.-M. Frahm, ``Structure-from-motion revisited,'' in \emph{Conference on Computer Vision and Pattern Recognition (CVPR)}, 2016.

\bibitem{shen2023anything}
Q.~Shen, X.~Yang, and X.~Wang, ``Anything-3d: Towards single-view anything reconstruction in the wild,'' \emph{arXiv preprint arXiv:2304.10261}, 2023.

\bibitem{su2022zebrapose}
Y.~Su, M.~Saleh, T.~Fetzer, J.~Rambach, N.~Navab, B.~Busam, D.~Stricker, and F.~Tombari, ``Zebrapose: Coarse to fine surface encoding for 6dof object pose estimation,'' in \emph{Proceedings of the IEEE/CVF Conference on Computer Vision and Pattern Recognition}, 2022, pp. 6738--6748.

\bibitem{thalhammer2023self}
S.~Thalhammer, J.-B. Weibel, M.~Vincze, and J.~Garcia-Rodriguez, ``Self-supervised vision transformers for 3d pose estimation of novel objects,'' \emph{arXiv preprint arXiv:2306.00129}, 2023.

\bibitem{wang2021gdr}
G.~Wang, F.~Manhardt, F.~Tombari, and X.~Ji, ``Gdr-net: Geometry-guided direct regression network for monocular 6d object pose estimation,'' in \emph{Proceedings of the IEEE/CVF Conference on Computer Vision and Pattern Recognition}, 2021, pp. 16\,611--16\,621.

\bibitem{wang2023visualgeometrygroundeddeep}
\BIBentryALTinterwordspacing
J.~Wang, N.~Karaev, C.~Rupprecht, and D.~Novotny, ``Vggsfm: Visual geometry grounded deep structure from motion,'' 2023. [Online]. Available: \url{https://arxiv.org/abs/2312.04563}
\BIBentrySTDinterwordspacing

\bibitem{Wang-SIG2022}
P.-S. Wang, Y.~Liu, and X.~Tong, ``Dual octree graph networks for learning adaptive volumetric shape representations,'' \emph{ACM Transactions on Graphics (SIGGRAPH)}, vol.~41, no.~4, 2022.

\bibitem{1284395}
Z.~Wang, A.~Bovik, H.~Sheikh, and E.~Simoncelli, ``Image quality assessment: from error visibility to structural similarity,'' \emph{IEEE Transactions on Image Processing}, vol.~13, no.~4, pp. 600--612, 2004.

\bibitem{point-cloud-utils}
F.~Williams, ``Point cloud utils,'' 2022, https://www.github.com/fwilliams/point-cloud-utils.

\bibitem{xiang2017posecnn}
Y.~Xiang, T.~Schmidt, V.~Narayanan, and D.~Fox, ``Posecnn: A convolutional neural network for 6d object pose estimation in cluttered scenes,'' \emph{arXiv preprint arXiv:1711.00199}, 2017.

\bibitem{xu2024instantmesh}
J.~Xu, W.~Cheng, Y.~Gao, X.~Wang, S.~Gao, and Y.~Shan, ``Instantmesh: Efficient 3d mesh generation from a single image with sparse-view large reconstruction models,'' \emph{arXiv preprint arXiv:2404.07191}, 2024.

\bibitem{zhang2018perceptual}
R.~Zhang, P.~Isola, A.~A. Efros, E.~Shechtman, and O.~Wang, ``The unreasonable effectiveness of deep features as a perceptual metric,'' in \emph{CVPR}, 2018.

\bibitem{zhang2024recognize}
Y.~Zhang, X.~Huang, J.~Ma, Z.~Li, Z.~Luo, Y.~Xie, Y.~Qin, T.~Luo, Y.~Li, S.~Liu, \emph{et~al.}, ``Recognize anything: A strong image tagging model,'' in \emph{Proceedings of the IEEE/CVF Conference on Computer Vision and Pattern Recognition}, 2024, pp. 1724--1732.

\end{thebibliography}
}

\end{document}